\documentclass{bmvc2k}

\title{Face Parsing via Recurrent Propagation}

\addauthor{Sifei Liu}{sliu32@ucmerced.edu}{1}
\addauthor{Jianping Shi}{shijianping@sensetime.com}{2}
\addauthor{Ji Liang}{liangji20040249@gmail.com}{3}
\addauthor{Ming-Hsuan Yang}{mhyang@ucmerced.edu}{1}

\addinstitution{
 University of California, Merced
}
\addinstitution{
 SenseTime, Inc.
}
\addinstitution{
 Momenta, Inc.
}

\runninghead{S. Liu, J. Shi, J. Liang, M.H. Yang}{Face Parsing via Recurrent Propagation}

\def\etal{\emph{et al}\bmvaOneDot}

\begin{document}

\maketitle

\begin{abstract}
Face parsing is an important problem in computer vision that finds numerous applications including recognition and editing.
Recently, deep convolutional neural networks (CNNs) have been applied to 
image parsing and segmentation with the state-of-the-art performance. 
In this paper, we propose a face parsing algorithm that combines 
hierarchical representations learned by a CNN, and accurate label propagations achieved by a spatially variant recurrent neural network (RNN).
The RNN-based propagation approach enables efficient inference 
over a global space 
with the guidance of semantic edges generated by a local convolutional model.
Since the convolutional architecture can be shallow and the spatial RNN can have few parameters, the framework is much faster and more light-weighted than the state-of-the-art 
CNNs for the same task.
We apply the proposed model to coarse-grained and fine-grained face parsing. 
For fine-grained face parsing, we develop a two-stage approach 
by first identifying the main regions and then segmenting the detail components, 
which achieves better performance in terms of accuracy and efficiency.
With a single GPU, the proposed algorithm parses face images accurately 
at 300 frames per second, which facilitates real-time applications.
\end{abstract}

\section{Introduction}
\label{sec:intro}
The recent years have witnessed significant progress in object segmentation and 
image parsing using deep CNNs~\cite{farabet2013,grangier2009,schulz2012,chen14semantic,crfasrnn_iccv2015,liu2015dpn}.
With end-to-end nonlinear classifiers and hierarchical features, CNN-based face parsing methods~\cite{liu2015multi,tsogkas2015semantic} achieve
the state-of-the-art performance than approaches based on hand-crafted features~\cite{GLOC_CVPR13}.
The main issues with existing CNN-based face parsing are the 
heavy computational load and large memory requirement. 
Both issues can be alleviated by using 
shallow or light-weighted convolutional structures, but at the expense of parsing accuracy. 

In this work, we propose a face parsing algorithm in which 
a spatially variant recurrent module is incorporated for global propagation of label information.
A straightforward combination of CNN and RNN is to take each activation in a CNN feature map as the input to a hidden recurrent node 
in a two-dimensional (2D) spatial sequence and use the recurrent structure to learn the propagation weight matrix in an end-to-end fashion~\cite{byeon2015scene,LiangSXFLY15}.
These models either utilize a spatial RNN~\cite{byeon2015scene}, 
or a stacked long short-term memory (LSTM)~\cite{LiangSXFLY15}. 
In contrast, the proposed recurrent structure exploits the strength of both models in which
we apply a simple structure similar to a typical RNN 
but maintains the capability of spatially variant propagation of an LSTM.
Specifically, the proposed model uses a spatially variant gate to control the propagation strength  over different locations in the label space.
For face parsing, this gate is naturally associated with the semantic boundary.
A gate allows propagation between pixels in a label-consistent region or stops it otherwise.
We show that this gate can be obtained via a relatively shallow CNN 
that focuses on learning low and mid-level representations. 
The RNN module, controlled by the gate, can utilize rich redundant information by propagating the predicted labels to their neighboring pixels in the label-consistent region.
Compared to a deep CNN face parser with similar performance, the propagation layer requires a small amount of model parameters and significantly reduces the computational cost.
As a result, we construct a model that is hundreds of times faster and smaller than 
deep CNN-based methods~\cite{liu2015multi,tsogkas2015semantic} for face parsing without loss of accuracy.

We validate the proposed algorithm on both 
coarse-grained (parsing an image with major regions including skin, hair and background) 
and fine-grained  (parsing an image with detailed facial components such as eyes, eyebrows, nose and mouth) face parsing. 
Both are of critical importance for real-world applications in face processing, 
e.g., coarse-grained face parsing for style transfer~\cite{Doodles16}
and fine-grained face parsing for virtual makeup.
Parsing only the main classes is generally easier under the same settings due to the complexity of solutions and more balanced distributions of training samples. 
%
%
We show that the proposed model can parse all faces of an image in one shot, 
and significantly outperform the state-of-the-art methods in terms of accuracy and speed.

One issue with applying a single network to fine-grained face parsing is 
the performance on small facial components.
This is due to the extremely unbalanced sample distributions and image size of these regions. 
We design a two-stage method to parse these components efficiently.
We train the model for the main classes in the first stage and then focus on the others 
with relatively simpler sub-networks.
Specifically, the sub-networks in the second stage take a cropped facial region as input.
In contrast to a face component may occupy a small amount of pixels with a whole image, the distributions of the pixels for a cropped region are more balanced.
We show that by dividing the second face parsing problem into several sub-tasks, 
the overall network complexity is significantly reduced.

The contributions of this work are summarized as follows. 
First, a light-weighted network is proposed for pixel-wise face parsing by combining a shallow CNN and a spatially variant RNN, which significantly reduces the computational load of deep CNN. 
Second, We show that when parsing a face image with multiple detailed components,  dividing the problem into several sub-tasks is significantly more efficient than using one single model, with even better accuracy. 
Experimental results on 
numerous datasets 
demonstrate the efficiency and effectiveness of the proposed face parsing algorithm against
the state-of-the-art methods.


\vspace{-2mm}
\section{Related Work}
\vspace{-2mm}

{\flushleft \bf Face Parsing.}
Face parsing considered in this work assigns dense semantic labels to all pixels in an image.
Typically, only one face image is assumed to be detected in an input frame.
Several approaches have been developed based on 
graphical models~\cite{GLOC_CVPR13}, exemplars~\cite{smith2013exemplar} and convolution networks~\cite{luo2012hierarchical,liu2015multi}.
The face parsing method developed by Luo \textit{et al.}~\cite{luo2012hierarchical}
hierarchically combines several separately trained deep models.
Liu \textit{et al.}~\cite{liu2015multi} develop a unified model to generate complete labels of facial regions in one single pipeline. 
In~\cite{Yamashita2015} Yamashita \textit{et al.} propose a weight cost function to deal with unbalanced samples for parsing small regions.  
Closest to this work is the multi-objective CNN \cite{liu2015multi} which introduces additional supervision of semantic edges and achieves substantial improvements for coarse-grained  and fine-grained face parsing.
However it is less effective for parsing detailed facial components and 
computationally expensive.
We show these issues can be largely resolved by the proposed methods.

{\flushleft \bf Recurrent Neural network.}
Recurrent networks~\cite{graves2013speech,Donahue_2015_CVPR,GregorDGW15,byeon2015scene,cho2014properties} 
have been shown to be effective for modeling long term dependencies in sequential data (e.g., speech). 
%
For image data, we can apply one-dimensional (1D) RNN to multiple dimensions in row/column-wise manner \cite{mdrnn2007,visin2015renet,KalchbrennerDG15}
or  multi-dimensional RNN (MDRNN)~\cite{mdrnn2007} such that each neural node can receive informations from multiple directions~\cite{byeon2015scene,oord2016pixel}
(as opposed to one direction in the conventional RNN).
In addition, there are other variants that leverage these two models, e.g., the grid LSTM~\cite{KalchbrennerDG15,LiangSXFLY15}.
The proposed model belongs to the first category, which is easier for parallelization.  

The proposed recurrent model is closely related to two recent image parsing methods~\cite{ChenBPMY15,liu2016learning} that 
utilize linear recurrent formulation to associate the adjacent pixels in either the semantic 
label space or the low-level image space.
In~\cite{ChenBPMY15} Chen~\etal propose the recurrent model with the concept of domain transform, where object edges learned on top of the intermediate layers of a fully convolutional network (FCN~\cite{long2014}) are used to regularize the transforms between adjacent pixels.
Liu \etal~\cite{liu2016learning} further extend the recurrent structure to high-order recursive filters to model more variations in the low-level space.
We compare different RNN frameworks with respect to the spatially invariant~\cite{zuo2015convolutional} and variant (as proposed in this work) recurrent modules.
%
In addition, we show that the formulation with recurrent propagation can substantially 
simplify the network structures with lower computational loads, 
which are not well exploited in the existing methods. 


\vspace{-3mm}
\section{Proposed Algorithm}
\label{sec:proposed_approach}
Most CNN-based face parsing algorithms~\cite{liu2015multi,Yamashita2015} apply deep networks with a large number of parameters, 
which entail heavy computational loads.
On the other hand, shallow models can be executed efficiently but not able to model global data dependency. 
%
%
In this work, we use a shallow CNN with a combination of spatially variant recurrent propagation module to model image data effectively and efficiently.
%

\begin{figure}[t!]
	\centering
	\includegraphics[width=0.9\textwidth]
	{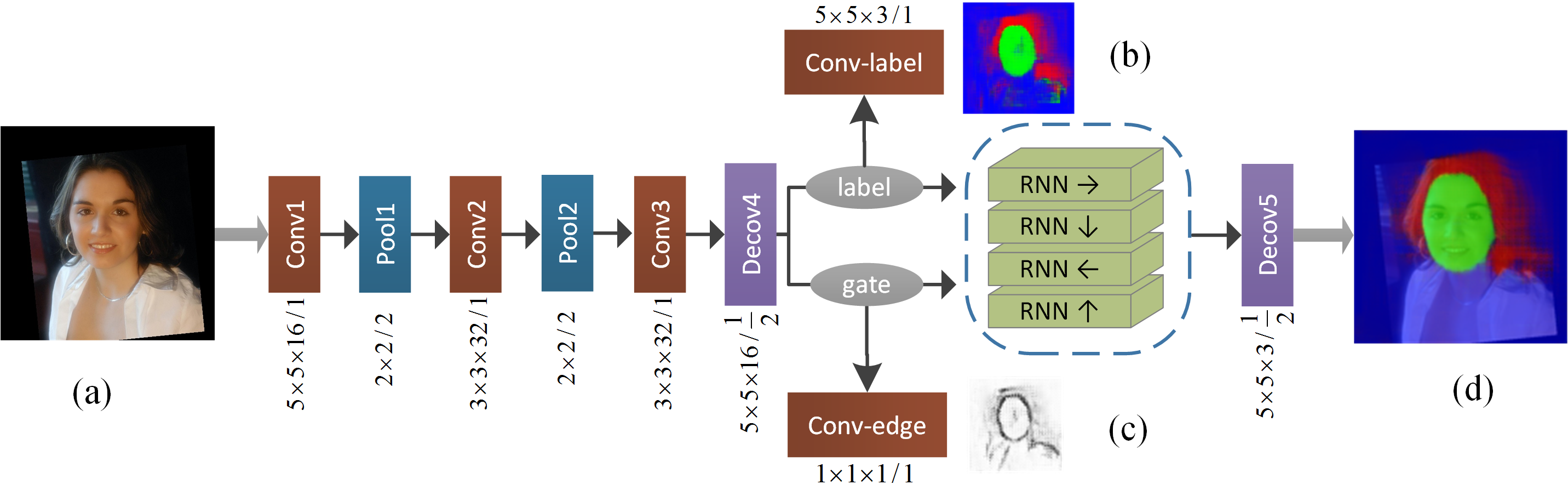}
	\caption{
\small Proposed parsing network architecture by combining a CNN and a spatial RNN.
		The CNN generates a coarse label map (b) and a recurrent gate (c), which are fed into 4 RNNs with different directions to generate a more accurate result (d).
		%
		The network structure is shown where 
		the notation for Conv1 ``5$\times$5$\times$16/1'' means convolution layer with $5\times5$ kernel, 16 channels and stride 1.
		The face image in (d) is further segmented with detailed labels in the second stage (see text and Figure~\ref{fig:stage2}).
	}
	\label{fig:stage1}
\end{figure}

%
Our model contains a shallow CNN and a spatial RNN, as shown in Figure~\ref{fig:stage1}.
First, the CNN takes a color image as its input and learns a coarse pixel-wise label score map (Figure~\ref{fig:stage1}(b)).
Second, the coarse label result is fed to a spatial recurrent unit for global propagation.
Specifically, the spatial propagation is controlled by a gate map  (Figure~\ref{fig:stage1}(c)), 
which is referred to as a recurrent gate in the rest of the paper. 
Each pixel in the map, formulated as a scalar weight coefficient to the recurrent term, controls the connection between two adjacent hidden nodes at the corresponding location.
Since a gate map can be supervised by the ground truth semantic boundaries from labeled annotations, it enables the recurrent propagation to be discriminative between semantically consistent and inconsistent regions, with respect to the specific input image.

We first briefly review conventional RNNs and describe how we extend it to the 2D space for image data, before introducing the recurrent gates.
We then discuss how to train the hybrid model in an end-to-end fashion.

\subsection{Recurrent Neural Networks}
\label{sec:RNN}
The conventional RNN is developed to process 1D sequential data where
each hidden node represents a single character, frame, pixel and is connected to its adjacent neighbor. 
The hidden node $i$, denoted as $h_i \in H$ receives two inputs: an external input $x_i \in X$ and its previous activation $h_{i-1}$ from one step back.
The summation of these two inputs is then non-linearly mapped via a function $\theta\left(\cdot\right)$ as the activation of the current step: 
\begin{eqnarray}
& h_i = \theta\left( a_i\right), \quad a_i = \omega_x x_i + \left(\omega_h h_{i-1}+b\right).
\label{eq:01}
\end{eqnarray}
In this formulation, $x_i$ and $h_i$ can have different dimensions, where the input transition matrix $\omega_x$ aligns $x_i$ to have the same dimension as $h_i$.
%
In addition, $b$ is a bias or offset
term to model data points centered at a point other than the origin.
For simplicity, we set $x_i$ and $h_i$ to have the same dimension, and remove the $\omega_x$ so that only the recurrent state transition matrix 
needs to be learned.

To extend the 1D RNN in~\eqref{eq:01} to 2D images, we consider each row/column as 1D sequence, and then adopt an approach similar to the bidirectional recurrent neural
network for processing temporal sequences~\cite{mdrnn2007}.
First, the 1D sequential RNN is processed respectively along left-to-right, top-to-bottom, and their reverse ways.
%
Taking the left-to-right direction for a 2D feature/label map as an example, 
the 1D sequential RNN scans each row from left to right.
As a result, four hidden activation maps are generated.

The four hidden activation maps can be grouped either in parallel or cascade, as introduced in~\cite{liu2016learning}.
The four maps share the same input $X$ with the parallel method, while in the cascade way, each RNN takes the output from a previous RNN as its input.
We adopt the parallel method and integrate the maps by selecting the optimal direction based on the
maximum response at each location.
This is carried out by a node-wise max pooling operation that can 
effectively select the maximally responded direction as the desired information to be propagated
and reject noisy information from other directions.
We note that the four-directional RNNs with parallel integration can be executed simultaneously with multiple GPUs for further acceleration as they are independent.

The backward pass is also an extension of 
the back propagation through time (BPTT) method used in RNNs~\cite{williams1995gradient}. 
Due to space limitation, we only present the derivative with respect to $a_i$:
\begin{equation}
\delta_i =\theta'\left( a_i\right)\cdot\left( \xi_i
+\omega_h\delta_{i+1}\right), 
\label{eq:03}
\end{equation}
where $\omega_h$ is a square weight matrix and all the others are 1D vectors.
We denote $\xi$ the influence from the output layer on top of the
proposed spatial RNN, 
and the second term in~\eqref{eq:03} the influence from the next
hidden node.
The derivatives are passed back in reverse order against the feedforward process, with four distinct directions computed respectively~\cite{williams1995gradient}. 

\subsection{Spatially Variant Recurrent Network}\label{sec:grn}
The fundamental problem of the RNN in~\eqref{eq:01} is that the hidden state transition matrix $\omega_h$ is spatially invariant.
%
As such, it tends to propagate any pixel to its adjacent ones with a group of fixed weights.
However, the label space is spatially variant with respect to different locations.
The propagation between pixels that share the same label
should be distinguished from those between pixels with different labels on the semantic boundaries. 

To this end, we propose a spatially variant recurrent network with gate maps $g_i\in G$ as an additional input to the spatial RNN. 
Each $g_i$ is an additional coefficient that controls the strength of connections between nodes to guide the recurrent propagations.
Intuitively, strong connections (e.g., $g_i$ is close to 1) should be enforced between nodes
in the label-consistent region.
On the other hand, weak connections (e.g., $g_i$ is close to 0) should be assigned to the nodes belonging to semantically different categories, so that they can be successfully separated.

To reformulate the framework, we have two types of inputs to a RNN, i.e., an external input $X$, and a spatially variant gate $G$.
Given a hidden node $h_i$, the spatially controllable recurrent propagation is:
\begin{eqnarray}
a_i = x_i + g_i\cdot\left(\omega_h h_{i-1}+b\right).
\label{eq:06}
\end{eqnarray}
The propagation of the hidden activation at $i-1$ to $i$ is controlled by dot product with $g_i$.
We use the identity function $\theta\left(x\right)=x $ as the activation (also used by~\cite{ChenBPMY15,liu2016learning}), since experimentally it achieves better performance.
To maintain the stability of the linearized formulation, the absolute value of  $g_i$, and norm of $\omega_h$ are both normalized to be within one during parameter update in order to prevent the hidden activation in $H$ to be increased exponentially. 

Similar to the sequential RNN, the BPTT algorithm is adopted to adjust $X$ and $G$ in the spatially variant RNN.
The derivatives with respect to $a_i$ and $g_i$, denoted as $\delta_i$ and $\varepsilon_i$ are:
\begin{equation}
\delta_i = \xi_i +g_i\cdot \omega\delta_{i+1},\qquad \varepsilon_i = \delta_i\cdot \left(\omega_h h_{i+1}+b\right).
\label{eq:07}
\end{equation}
In addition, the derivative from RNN with respect to $x_i$ is equal to $\delta_i$.

\subsection{Hybrid Model of CNN and RNN}
\label{sec:mulit}
In the proposed framework, we apply a CNN that provides label representation $X$ and spatially variant gate representation $G$ to the spatial RNN
(see Figure~\ref{fig:stage1}). 
With the effective propagation of RNN, the CNN can be relatively shallow as 
revealed in the experimental analyses.
Taking the three-class face parsing as an example, the main part of CNN is equipped with only three convolutional layers, two max pooling (down-sampling) as well as 
deconvolutional  (upsampling) layers, as shown in Figure~\ref{fig:stage1},
and at most 32 channels for each layer.
%
%
The proposed network is significantly smaller than most existing CNN-based face parsing models based on 6 convolutional layers with 2 fully-connected layers~\cite{liu2015multi}, 
or 16 layers~\cite{tsogkas2015semantic} from VGG~\cite{simonyan2014very}.

To connect with the spatial RNN, the feature maps with 16 channels generated from the first deconvolutional layer (Deconv6 in Figure~\ref{fig:stage1}) 
are equally split into two components (each with 8 channels), 
where one is for pixel-wise labels and the other is for the recurrent gate, with equal width and height.
They are then fed to four recurrent layers with different directions as $X$ and $G$ respectively, where each pixel $i$ in the hidden layers is processed by combining $x_i$ and $g_i$ based on~\eqref{eq:06}.

The hybrid network contains three different loss layers.
At the top of the CNN, both $X$ and $G$ are supervised with the softmax cross entropy loss.
The labeling representations are transferred by a convolutional layer to be directly supervised by the ground truth labels (see Figure~\ref{fig:stage1}(b)).
The gate representations are transferred by a $1\times 1$ convolutional layer to have 
a single channel output, which is supervised by the boundary between different categories (see Figure~\ref{fig:stage1}(c)).
Finally, the output of RNN with 8 channels are transferred to 3 channels, upsampled to the original image scale, and supervised by the ground truth labels (see Figure~\ref{fig:stage1}(d)).
All the losses encourage the CNN to learn better label candidates as well as guidances to the propagation.
Specifically, the ground truth boundaries are obtained from the annotated labels, by setting its boundary pixels to zeros and all the others to one.
For example, with a pixel $i$ that is located on a boundary of two categories, the ground truth value is set to zero, which can encourage the $g_i$ to ``cut off'' the connection between different classes, and vise-versa.


\section{Sub-networks for the Detailed Components}
As discussed in Section~\ref{sec:intro} and revealed in the experiments, a single network does not perform well on small facial components.
One problem is that some facial components amount to small percentages of the entire dataset, e.g., the eye regions in Figure~\ref{fig:stage1}(a) 
occupy less than $1\%$ of the whole image.
It is difficult to parse such components in one stage due to unbalanced labeled data.
The work of~\cite{liu2015multi} applies a simple strategy by sampling with an equal number of input patches.  
However, the performance on small facial components is not satisfactory compared to categories with more pixels, e.g., skin.
The other problem is the limited resolution of facial components.
With a larger input image, more details of the components can be learned.
However, it requires deeper or larger models to adapt to the enlarged receptive fields.
For a single model, it is a trade-off between effectiveness and efficiency.

We decompose a unified face segmentation network into a two-stage framework.
In practice, parsing major classes with either frontal, canonical face or multiple random faces
can be handled using the first stage only.
For parsing 11 classes in the HELEN dataset, each component can be labeled independently 
first and then combined with the major ones. 

{\flushleft \bf First Stage Face-Hair-Background Network.}
The first stage network classifies an image into skin, hair and background regions 
using the combination of CNN and RNN, as introduced in Section~\ref{sec:proposed_approach}. 
Since there are only three labels with relatively equal distribution, 
we do not need to balance the samples. 
As these classes do not contain detailed structures such as facial components, the input resolution does not need to be high.
Similar to~\cite{liu2015multi}, a face image is detected and roughly aligned to the center using~\cite{sun2013deep}, with a resolution of $128\times 128$ pixels. 
The result of the label has the same resolution as the input image.

\begin{figure}[t!]
	\centering
	\includegraphics[width=0.9\textwidth]
	{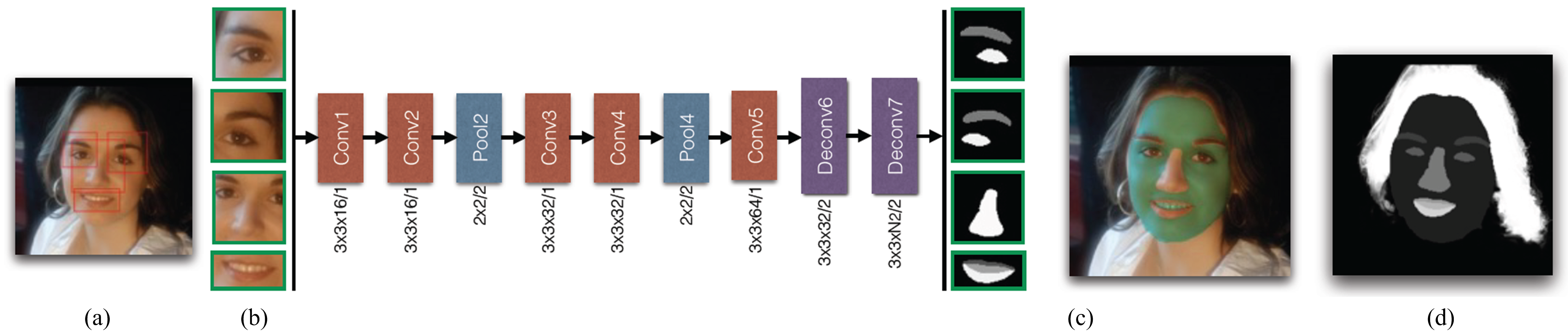}
	\caption{
\small
The second stage network operates on the cropped region, i.e., 
left and right eyes, nose, and mouth, to parse accurate facial components.
		The final parsing result in (d) is the combination of segments from two stages. 
	}
	\label{fig:stage2}

\end{figure}

{\flushleft \bf Second Stage Facial Component Networks.}
We locate the facial components for high resolution faces image through 5 detected key points (eye centers, nose tip, and mouth corners)~\cite{sun2013deep}, 
and crop the patches accordingly. 
We train three simple and efficient networks to segment eye and eyebrow, nose, and 
mouth regions, respectively.
Figure~\ref{fig:stage2}(b) shows the structure of eye/eyebrow network. 
It contains five convolution layers, two max-pooling layers, and two deconvolution layers, with an input size of $64\times 64$. 
Similar network structures are used for the nose as well as the mouth, and the input image size 
is $64\times 64$ and $32\times 64$, respectively. 
Since each image is cropped around each facial component, it does not include many pixels from the skin region. 
Therefore, the sample distribution is balanced for network training. 
The final parsing result is composed of the accurate facial component segments in the second stage and 
the coarse segments in the first stage.  
Since the segmentation task in the second stage is easier, we do not apply the component-wise spatial RNN for efficiency reason.



\section{Experimental Results}\label{sec:exp}
We carry out experiments on images containing one or multiple faces. 
For single face parsing, we evaluate our method on 
the LFW-PL~\cite{GLOC_CVPR13} and HELEN~\cite{smith2013exemplar}  datasets.
In addition, we develop a Multi-Face dataset to evaluate parsing numerous faces in one image.
All experiments are conducted on a Nvidia GeForce GTX TITAN X GPU.

\subsection{Datasets and Settings}

{\flushleft \bf LFW-PL and HELEN Datasets.}
The LFW part label (LFW-PL) dataset contains $2,927$ face images. 
Each face image is annotated as skin, hair or background using superpixels, and  
roughly aligned to the center~\cite{GLOC_CVPR13}.
The HELEN dataset contains $2,330$ face images with manually labeled facial components including eyes, eyebrows, nose, lips, etc. 
For both datasets, the most centered face in each image is annotated. 
We adopt the same setting of data splits as~\cite{liu2015multi} and resize each image and its corresponding label to $128\times 128$. 
For the HELEN, dataset, the hair region is trained as one category in the first stage of our algorithm but is not evaluated for fair comparisons with the existing method~\cite{smith2013exemplar,liu2015multi}.

\begin{figure}[t!]
	\centering
	\begin{tabular}{@{\hspace{0mm}}c@{\hspace{0.5mm}}c@{\hspace{0.5mm}}c@{\hspace{0.5mm}}c@{\hspace{0.5mm}}c@{\hspace{0.5mm}}c}
		\includegraphics[width=0.15\linewidth]{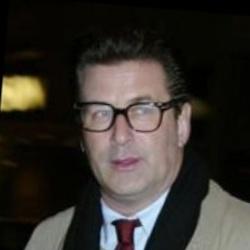} &
		\includegraphics[width=0.15\linewidth]{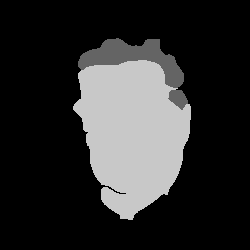} &
		\includegraphics[width=0.15\linewidth]{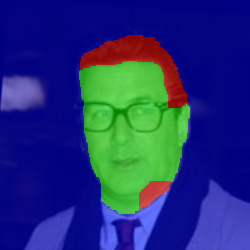} &
		\includegraphics[width=0.15\linewidth]{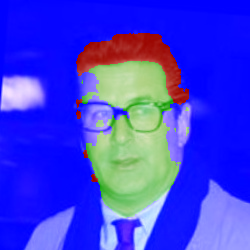} &
		\includegraphics[width=0.15\linewidth]{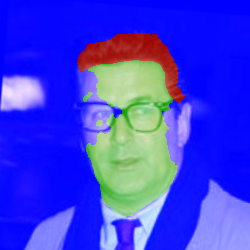} &
		\includegraphics[width=0.15\linewidth]{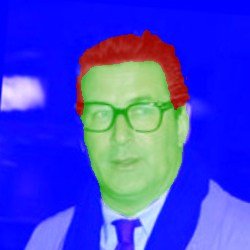} \\
		{\footnotesize (a) } &{\footnotesize (b) }  &{\footnotesize (c) }  &{\footnotesize (d) }  &{\footnotesize (e) } & {\footnotesize (f) } \\
	\end{tabular}
	\caption{
\small
		Face parsing results on the LFW-PL dataset. 
		(a) input image. 
		(b) ground-truth annotations. 
		(c) results from~\cite{liu2015multi}. 
		(d) results from CNN-S. 
		(e) results from CNN with dense CRF. 
		(f) results by RNN-G. 
		More results are presented in the supplementary material.}
	\label{fig:LFW-PL}\vspace{-3mm}
\end{figure}

\begin{figure}[t!]
	\centering
	\begin{tabular}{@{\hspace{0mm}}c@{\hspace{0.5mm}}c@{\hspace{0.5mm}}c@{\hspace{0.5mm}}c@{\hspace{0.5mm}}c@{\hspace{0.5mm}}c@{\hspace{0.5mm}}c}
		\includegraphics[width=0.15\linewidth]{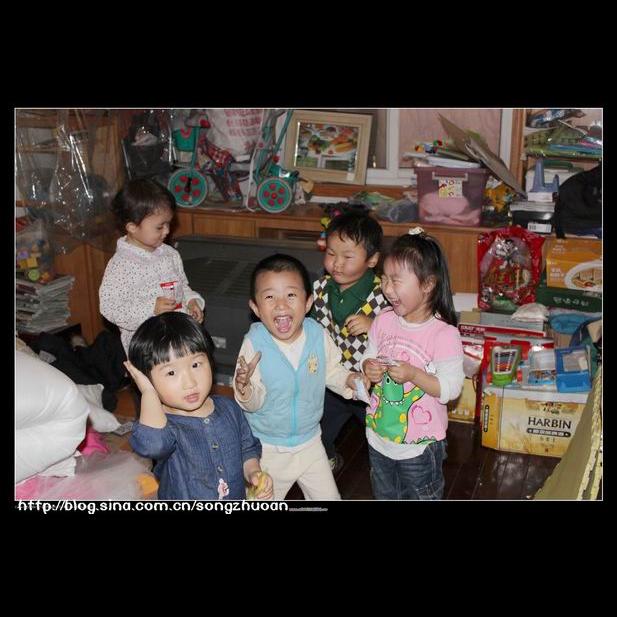} &
		\includegraphics[width=0.15\linewidth]{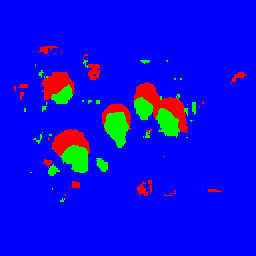} &
		\includegraphics[width=0.15\linewidth]{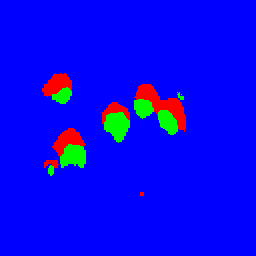} &
		\includegraphics[width=0.15\linewidth]{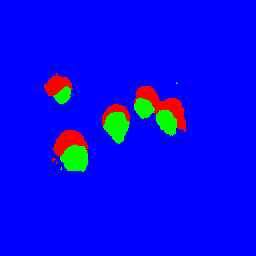} &
		\includegraphics[width=0.15\linewidth]{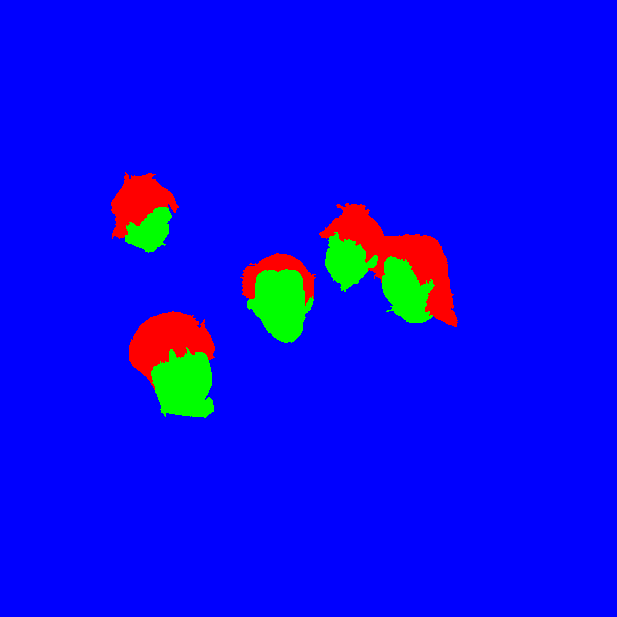} &
		\includegraphics[width=0.15\linewidth]{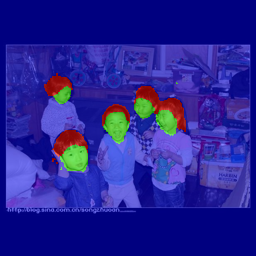} &\\
		{\footnotesize (a) } & {\footnotesize (b) }& {\footnotesize (c) } & {\footnotesize (d) } & {\footnotesize (e) } & {\footnotesize (f) }
	\end{tabular}
	\caption{
\small
		Parsing results on the Multi-Face dataset. 
		(a) input image. 
		(b) results by the baseline CNN. 
		(c) results by the standard RNN.
		(d) results from RNN-G. 
		(e) the ground truth.
		(f) a visualized version of RNN-G.
		Our method is able to effectively and efficiently parse multiple faces in the cluttered background.
		 More results are presented in the supplementary material.}
	\label{fig:multiface1}
\end{figure}

\begin{table}[t]
	\caption {
\small Quantitative results on the LFW-PL dataset.``F'' denotes f-score, ``AC'' denotes accuracy, ``bg'' denotes background, ``-'' denotes not available.}
	\label{tab:LFW-PL}
	{\footnotesize
		\begin{tabular}{ l c |c  c ||c  c c c c c }
			& \footnotesize{(\%)} &\scriptsize{GLOC~\cite{GLOC_CVPR13}} &\scriptsize{MO~\cite{liu2015multi}} &\scriptsize{CNN-S} & \scriptsize{CNN-deep}& \scriptsize{CNN-CRF~\cite{chen14semantic}} &\scriptsize{RNN~\cite{zuo2015convolutional}} &\scriptsize{RNN-G}  \\\hline
			&\footnotesize{F-skin} & - &\footnotesize{93.93} &\footnotesize{90.47} &\footnotesize{91.63} & \footnotesize{91.25} & \footnotesize{93.72} &\footnotesize{\textbf{97.55}} \\ 
			&\footnotesize{F-hair} & - &\footnotesize{80.70} &\footnotesize{76.09} &\footnotesize{78.30} & \footnotesize{75.21} & \footnotesize{81.21} & \footnotesize{\textbf{83.43}} \\
			&\footnotesize{F-bg} & - &\footnotesize{97.10} &\footnotesize{95.42} &\footnotesize{95.95} & \footnotesize{99.58} & \footnotesize{97.15} & \footnotesize{\textbf{94.37}} \\
			& \footnotesize{AC} &\footnotesize{94.95} &\footnotesize{95.09} &\footnotesize{92.44} &\footnotesize{93.27}& \footnotesize{92.59} &\footnotesize{94.85} &\footnotesize{\textbf{95.46}}  \\\hline
			& \footnotesize{Time (ms)} &\footnotesize{254 (CPU)} &\footnotesize{$\sim110$} &\footnotesize{$<1$} &\footnotesize{$\sim2$}& \footnotesize{$\sim7$} &\footnotesize{$\sim2$} &\footnotesize{$\sim2$}  \\\hline
		\end{tabular}
	}
\end{table}

\begin{table}[t]
	\centering
	\caption {
\small
	Quantitative results on the Multi-Face dataset.}
	\label{tab:multi}
	{
\footnotesize
		\begin{tabular}{ l c |c  c  c c c c }
			& \footnotesize{(\%)} & \footnotesize{CNN-deep}& \footnotesize{CNN-CRF} &\footnotesize{RNN} & \footnotesize{Det+RNN-G single}&\footnotesize{RNN-G}  \\\hline
			&\footnotesize{F-skin} & \footnotesize{75.56} & \footnotesize{77.84} & \footnotesize{73.33} &\footnotesize{81.02}& \footnotesize{\textbf{87.36}} \\ 
			&\footnotesize{F-hair} & \footnotesize{64.62} & \footnotesize{61.53} & \footnotesize{62.85} &\footnotesize{55.35}& \footnotesize{\textbf{73.09}} \\
			&\footnotesize{F-background} & \footnotesize{96.5} & \footnotesize{97.08} & \footnotesize{96.18} &\footnotesize{97.10}& \footnotesize{\textbf{98.19}} \\
			& \footnotesize{AC} & \footnotesize{93.39}& \footnotesize{94.5} &\footnotesize{92.78} &\footnotesize{94.42}&\footnotesize{\textbf{96.35}}  \\\hline
		\end{tabular}
	}
\end{table}

{\flushleft \bf Multi-Face Dataset.}
We collect a Multi-Face dataset where each image contains multiple faces.
It contains $9,645$ images in unconstrained environments with pixel-wise labels including skin, hair, and background. 
%
%
This dataset is divided into a training set of $9,045$ images, a
test set of $200$ images,  and a validation set of $200$ images.
We rescale each image and its
corresponding label according to the length of the long side to maintain the aspect ratio.
Each one is zero padded to result in a $512\times 512$ image where all faces appear clearly. 

{\flushleft \bf Network Implementation.}
Our network structures are described in Figure~\ref{fig:stage1} and~\ref{fig:stage2}.
We use the first stage model (see Figure~\ref{fig:stage1}) to parse images in the LFW-PL and Multi-Face datasets, and the facial skin and hair regions in the Helen dataset.
In addition, we use the second stage model (see Figure~\ref{fig:stage2}) to parse 
facial components of images in the HELEN dataset.

For fair comparison with the previous work, we align the input images according to~\cite{liu2015multi} in the HELEN dataset.
The faces in the LFW-PL dataset do not need additional processing since the released images are already coarsely aligned.
On the other hand, we directly use the $512\times 512$ images as the network inputs, and do not preprocess any face for the Multi-Face dataset.
We quantitatively evaluate and compare our model using per-pixel accuracy and F-measure for each class in all experiments.

In the first stage, the boundaries in Figure~\ref{fig:stage1}(c) are balanced with the ratio of positive/negative number of pixels set to $1:5$ such that a sufficient number of boundary samples can be drawn.
The training images augmented by random affine and mirror transformations
for increasing the variation of training samples.
The network for the Multi-Face dataset has two more $3\times 3\times 16$ convolutional units (with max-pooling) and one more deconvolutional layer to adapt to the input size. 
The results are evaluated with the resolution of $256 \times 256$.
The boundary loss sampling and training image augmentation strategies are uniformly applied to all experiments.
For the second stage model, we crop the facial components based on the 5 facial key points from~\cite{sun2013deep} for training and tests.  
We include at least additional 20\% height/width of the total foreground height/width in the cropped images during training and maintain the aspect ratio.


\subsection{Coarse-grained Face Parsing}
Face parsing with 3 classes are carried out using the first stage model
on the LFW-PL and the Multi-Face datasets, respectively. 
We compare the proposed method, denoted as RNN-G with: 
(a) shallow CNN part only (CNN-S). (b) shallow CNN with the RNN module replaced by two $3\times3$ convolutional layers with 32 channels as a baseline network, denoted as CNN-Deep.
We increase the number of the output channels of Deconv6 (Figure~\ref{fig:stage1}) from $8$ to $16$ to ensure that the shallow model can converge.
(c) a combination of the shallow CNN and the post processing with a dense CRF, denoted as CNN-CRF, which is commonly used in 
recent semantic segmentation tasks~\cite{chen14semantic}. 
(d) a standard RNN in~\eqref{eq:01} (similar to~\cite{zuo2015convolutional}) with the same CNN.
We note that both~\cite{chen14semantic,zuo2015convolutional} do note have experiments on shallow networks.

We show two more baseline methods~\cite{GLOC_CVPR13,liu2015multi} evaluated on 
the LFW-PL dataset.
%
Specifically, we adjust~\cite{liu2015multi} by using only one-time feedforward with $2\times$ bilinear upsampling layer for fair comparisons in speed and accuracy.
For the Multi-Face dataset, we use the models to parse all faces in images without using any detector.
This is computationally efficient and useful 
for numerous applications without the need of  instance-level information.
We note the label distribution of the Multi-Face dataset 
with respect to different categories are significantly unbalanced since the vast 
majority of pixels belong to the background regions.
Thus, we apply a data sampling strategy at each loss layer by maintaining the number of sampled background pixels as 5 times of the total number of pixels for skin and hair regions. 

\begin{figure*}[t!]
	\begin{tabular}{@{\hspace{0mm}}c@{\hspace{0mm}}c@{\hspace{0mm}}c@{\hspace{0mm}}c@{\hspace{1.5mm}}c@{\hspace{0mm}}c@{\hspace{0mm}}c@{\hspace{0mm}}c@{\hspace{0mm}}c}
		\includegraphics[width=0.124\linewidth]{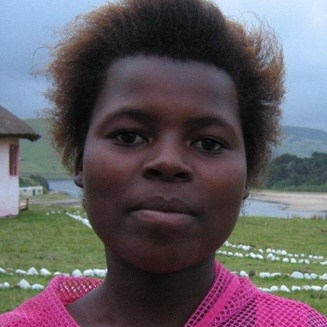} &
		\includegraphics[width=0.124\linewidth]{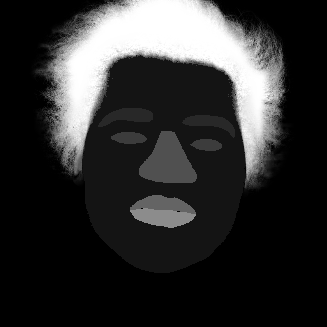} &		
		\includegraphics[width=0.124\linewidth]{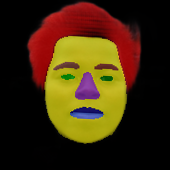} &
		\includegraphics[width=0.124\linewidth]{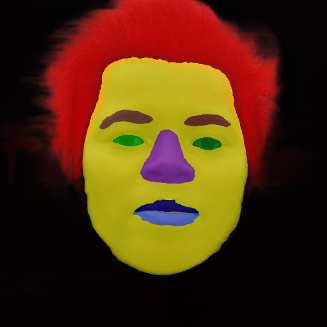} &		
		\includegraphics[width=0.124\linewidth]{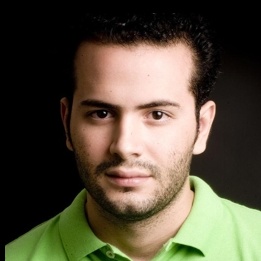} &
		\includegraphics[width=0.124\linewidth]{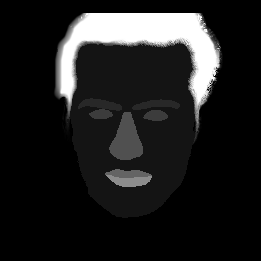} &
		\includegraphics[width=0.124\linewidth]{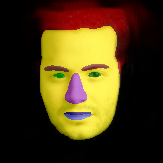} &
		\includegraphics[width=0.124\linewidth]{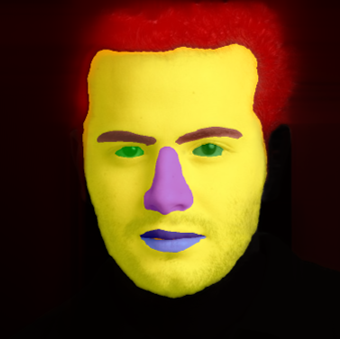}\\
		{\footnotesize (a) } & {\footnotesize (b) }& {\footnotesize (c) } & {\footnotesize (d) } & {\footnotesize (a) } &{\footnotesize (b) } &{\footnotesize (c) } &{\footnotesize (d) }
	\end{tabular}
	\caption{
\small
		Face parsing results on the HELEN~\cite{smith2013exemplar} dataset.
		(a) input image.
		(b) ground-truth annotations. 
		(c) results from~\cite{liu2015multi}. 
		(We roughly crop the results for better visual comparisons.)
		(d) our results with 11-class pixel-wise parsing. More results are presented in the supplementary material.
		}
	\label{fig:HELEN}\vspace{-3mm}
\end{figure*}

\begin{table*}[t]
	\centering
	\caption{
\small Quantitative evaluation results on the HELEN dataset. We denote the upper and lower lips as ``U-lip'' and ``L-lip'', and overall mouth part as ``mouth'', respectively. See~\cite{liu2015multi,smith2013exemplar} for more details.}
	\footnotesize
	\begin{tabular}{c|*{8}{c}r}
		{ Methods }             & eyes & brows & nose & in mouth & U-lip  & L-lip & mouth &  skin & overall \\
		\hline
		{ Liu~\etal~\cite{liu2011nonparametric}} & 77.0 & 64.0 & 84.3 & 60.1 & 65.0 & 61.8 & 74.2 & 88.6 & 73.8\\
		{ Smith~\etal~\cite{smith2013exemplar}} & 78.5 & 72.2 & 92.2 & 71.3 & 65.1 & 70.0 & 85.7 & 88.2 & 80.4  \\
		{ Liu~\etal~\cite{liu2015multi}}           & 76.8 & 71.3 & 90.9 & 80.8 & 62.3 & 69.4 & 84.1 & 91.0 & 84.7  \\
		\hline
		{ Ours 1-stage }    & {63.3} & {53.7} & {87.5} & {65.7} &  {54.0} & {72.6} &  {80.6} & {91.1} & {78.8}  \\
		{ Ours 2-stage}    & \bf{86.8} & \bf{77.0} & \bf{93.0} & {79.2} &  \bf{74.3} & \bf{81.7} &  \bf{89.1} & \bf{92.1} & \bf{88.6}  \\\hline
	\end{tabular}
	\label{tab:helen}
\end{table*}

Table~\ref{tab:LFW-PL} and \ref{tab:multi} show the results with similar trends on 
these two datasets.
Overall, the shallow CNN, i.e.,  {CNN-S}, has limited performance. 
There is no significant improvement gain by simply adding more layers (CNN-Deep) or adding an additional dense CRF module (CNN-CRF).
The standard RNN without the spatially variant gate performs better, but still worse than the proposed method. 
With the spatially variant gate, the RNN-G model performs significantly better than the baseline {CNN-S}, {CNN-Deep} and RNN models. 
The results demonstrate the effectiveness of the proposed spatially variant RNN structure. 
The proposed models operate at $500$ fps for a $128\times 128$ single face image 
and $200$ fps for a $512\times 512$ image with multiple faces.

Figure~\ref{fig:LFW-PL} and~\ref{fig:multiface1} show some parsing results 
on the two datasets.
The proposed RNN-G model performs favorably against the CNN-S, CNN-CRF, standard RNN, and the method using nonparametric prior and graph cut inference~\cite{liu2015multi}. 
%
%
For Multi-Face dataset, we evaluate the alternative method using a face detector~\cite{qin2016joint} and the 
single face parser trained on the LFW-PL dataset, which operates at $37$ fps on average (depending on the number of detected faces). 
The RNN-G model performs favorably in the cluttered background against all alternative methods in terms of accuracy and efficiency. 

\vspace{-3mm}
\subsection{Fine-grained Face Parsing}
In the HELEN dataset, we evaluate the parsing  results following the settings in~\cite{liu2015multi}, where
the second stage network is utilized to improve parsing results.
Since the second stage takes less than 1 ms, the overall run-time for parsing a face with 11 classes can operate at $300$ fps on a single GPU.

Table~\ref{tab:helen} and Figure~\ref{fig:HELEN} show the quantitative and qualitative parsing results. 
We first show that by using a single stage, the unified model cannot handle detailed facial parts even with the spatially variant RNN module.
%
Our two-stage network performs favorably against the state-of-the-art methods, 
and the one stage network model, on all categories.
It is worth noting that the overall F-measure achieved by the RNN-G model is 0.886, which 
amounts to 20\% reduction in error rate from the state-of-the-art method~\cite{liu2011nonparametric}.
%
These experimental results demonstrate that the two-stage network structure with the spatially variant gate is effective for accurate and efficient face parsing.

\section{Conclusions}
In this paper, we propose a pixel-level face parsing network by combining a shallow CNN and a spatially variant RNN.
The recurrent propagation infers globally over the entire image with the guidance of a local model, which 
reduces the computational load of deep CNNs.
We develop a two-stage approach for accurate parsing of the detailed facial component. 
Experimental results on the HELEN~\cite{smith2013exemplar}, LFW-PL~\cite{GLOC_CVPR13} and the proposed Multi-Face datasets 
demonstrate the efficiency and effectiveness of the proposed face parsing algorithm against 
the state-of-the-art methods. 

\paragraph{Acknowledgment}
This work is supported in part by SenseTime Inc. and the NSF CAREER Grant \#1149783, gifts from NEC and Nvidia.

{\small
	\bibliographystyle{ieee}
	\bibliography{cnn-rnn}

\begin{thebibliography}{32}
\providecommand{\natexlab}[1]{#1}
\providecommand{\url}[1]{\texttt{#1}}
\expandafter\ifx\csname urlstyle\endcsname\relax
  \providecommand{\doi}[1]{doi: #1}\else
  \providecommand{\doi}{doi: \begingroup \urlstyle{rm}\Url}\fi

\bibitem[Byeon et~al.(2015)Byeon, Breuel, Raue, and Liwicki]{byeon2015scene}
Wonmin Byeon, Thomas~M Breuel, Federico Raue, and Marcus Liwicki.
\newblock Scene labeling with lstm recurrent neural networks.
\newblock In \emph{CVPR}, 2015.

\bibitem[Champandard(2016)]{Doodles16}
Alex~J. Champandard.
\newblock Semantic style transfer and turning two-bit doodles into fine
  artworks.
\newblock \emph{arXiv preprint arXiv:1603.01768}, 2016.

\bibitem[Chen et~al.(2015{\natexlab{a}})Chen, Barron, Papandreou, Murphy, and
  Yuille]{ChenBPMY15}
Liang{-}Chieh Chen, Jonathan~T. Barron, George Papandreou, Kevin Murphy, and
  Alan~L. Yuille.
\newblock Semantic image segmentation with task-specific edge detection using
  cnns and a discriminatively trained domain transform.
\newblock \emph{arXiv preprint arXiv:1511.03328}, 2015{\natexlab{a}}.

\bibitem[Chen et~al.(2015{\natexlab{b}})Chen, Papandreou, Kokkinos, Murphy, and
  Yuille]{chen14semantic}
Liang-Chieh Chen, George Papandreou, Iasonas Kokkinos, Kevin Murphy, and Alan~L
  Yuille.
\newblock Semantic image segmentation with deep convolutional nets and fully
  connected crfs.
\newblock In \emph{ICLR}, 2015{\natexlab{b}}.

\bibitem[Cho et~al.(2014)Cho, van Merri{\"e}nboer, Bahdanau, and
  Bengio]{cho2014properties}
Kyunghyun Cho, Bart van Merri{\"e}nboer, Dzmitry Bahdanau, and Yoshua Bengio.
\newblock On the properties of neural machine translation: Encoder-decoder
  approaches.
\newblock \emph{arXiv preprint arXiv:1409.1259}, 2014.

\bibitem[David et~al.(2009)David, L{\'e}on, and Ronan]{grangier2009}
Grangier David, Bottou L{\'e}on, and Collobert Ronan.
\newblock Deep convolutional networks for scene parsing.
\newblock In \emph{ICML Deep Learning Workshop}, 2009.

\bibitem[Donahue et~al.(2015)Donahue, Hendricks, Guadarrama, Rohrbach,
  Venugopalan, Saenko, and Darrell]{Donahue_2015_CVPR}
Jeffrey Donahue, Lisa~Anne Hendricks, Sergio Guadarrama, Marcus Rohrbach,
  Subhashini Venugopalan, Kate Saenko, and Trevor Darrell.
\newblock Long-term recurrent convolutional networks for visual recognition and
  description.
\newblock In \emph{CVPR}, 2015.

\bibitem[Farabet et~al.(2013)Farabet, Couprie, Najman, and LeCun]{farabet2013}
Cl{\'e}ment Farabet, Camille Couprie, Laurent Najman, and Yann LeCun.
\newblock Learning hierarchical features for scene labeling.
\newblock \emph{IEEE PAMI}, 35\penalty0 (8):\penalty0 1915--1929, 2013.

\bibitem[Graves et~al.(2013)Graves, rahman Mohamed, and
  Hinton]{graves2013speech}
Alan Graves, Abdel rahman Mohamed, and Geoffrey Hinton.
\newblock Speech recognition with deep recurrent neural networks.
\newblock In \emph{ICASSP}, pages 6645--6649. IEEE, 2013.

\bibitem[Graves et~al.(2007)Graves, Fernández, and Schmidhuber]{mdrnn2007}
Alex Graves, Santiago Fernández, and Jürgen Schmidhuber.
\newblock Multi-dimensional recurrent neural networks.
\newblock In \emph{ICANN}, pages 549--558, 2007.

\bibitem[Gregor et~al.(2015)Gregor, Danihelka, Graves, and
  Wierstra]{GregorDGW15}
Karol Gregor, Ivo Danihelka, Alex Graves, and Daan Wierstra.
\newblock {DRAW:} {A} recurrent neural network for image generation.
\newblock \emph{arXiv preprint arXiv:1502.04623}, 2015.

\bibitem[Kae et~al.(2013)Kae, Sohn, Lee, and Learned-Miller]{GLOC_CVPR13}
Andrew Kae, Kihyuk Sohn, Honglak Lee, and Erik Learned-Miller.
\newblock Augmenting {CRF}s with {B}oltzmann machine shape priors for image
  labeling.
\newblock In \emph{CVPR}, 2013.

\bibitem[Kalchbrenner et~al.(2015)Kalchbrenner, Danihelka, and
  Graves]{KalchbrennerDG15}
Nal Kalchbrenner, Ivo Danihelka, and Alex Graves.
\newblock Grid long short-term memory.
\newblock \emph{arXiv preprint arXiv:1507.01526}, 2015.

\bibitem[Liang et~al.(2015)Liang, Shen, Xiang, Feng, Lin, and
  Yan]{LiangSXFLY15}
Xiaodan Liang, Xiaohui Shen, Donglai Xiang, Jiashi Feng, Liang Lin, and
  Shuicheng Yan.
\newblock Semantic object parsing with local-global long short-term memory.
\newblock \emph{arXiv preprint arXiv:1511.04510}, 2015.

\bibitem[Liu et~al.(2011)Liu, Yuen, and Torralba]{liu2011nonparametric}
Ce~Liu, Jenny Yuen, and Antonio Torralba.
\newblock Nonparametric scene parsing via label transfer.
\newblock \emph{PAMI}, 2011.

\bibitem[Liu et~al.(2015{\natexlab{a}})Liu, Yang, Huang, and
  Yang]{liu2015multi}
Sifei Liu, Jimei Yang, Chang Huang, and Ming-Hsuan Yang.
\newblock Multi-objective convolutional learning for face labeling.
\newblock In \emph{CVPR}, 2015{\natexlab{a}}.

\bibitem[Liu et~al.(2016)Liu, Pan, and Yang]{liu2016learning}
Sifei Liu, Jinshan Pan, and Ming-Hsuan Yang.
\newblock Learning recursive filters for low-level vision via a hybrid neural
  network.
\newblock In \emph{ECCV}, 2016.

\bibitem[Liu et~al.(2015{\natexlab{b}})Liu, Li, Luo, Loy, and Tang]{liu2015dpn}
Ziwei Liu, Xiaoxiao Li, Ping Luo, Chen~Change Loy, and Xiaoou Tang.
\newblock Semantic image segmentation via deep parsing network.
\newblock In \emph{ICCV}, 2015{\natexlab{b}}.

\bibitem[Long et~al.(2014)Long, Shelhamer, and Darrell]{long2014}
Jonathan Long, Evan Shelhamer, and Trevor Darrell.
\newblock Fully convolutional networks for semantic segmentation.
\newblock \emph{arXiv preprint arXiv:1411.4038}, 2014.

\bibitem[Luo et~al.(2012)Luo, Wang, and Tang]{luo2012hierarchical}
Ping Luo, Xiaogang Wang, and Xiaoou Tang.
\newblock Hierarchical face parsing via deep learning.
\newblock In \emph{CVPR}. IEEE, 2012.

\bibitem[Qin et~al.(2016)Qin, Yan, Li, and Hu]{qin2016joint}
Hongwei Qin, Junjie Yan, Xiu Li, and Xiaolin Hu.
\newblock Joint training of cascaded cnn for face detection.
\newblock In \emph{CVPR}, 2016.

\bibitem[Schulz and Behnke(2012)]{schulz2012}
Hannes Schulz and Sven Behnke.
\newblock Learning object-class segmentation with convolutional neural
  networks.
\newblock In \emph{ESANN}, volume~3, page~1, 2012.

\bibitem[Simonyan and Zisserman(2014)]{simonyan2014very}
Karen Simonyan and Andrew Zisserman.
\newblock Very deep convolutional networks for large-scale image recognition.
\newblock \emph{arXiv preprint arXiv:1409.1556}, 2014.

\bibitem[Smith et~al.(2013)Smith, Zhang, Brandt, Lin, and
  Yang]{smith2013exemplar}
Brandon Smith, Li~Zhang, Jonathan Brandt, Zhe Lin, and Jianchao Yang.
\newblock Exemplar-based face parsing.
\newblock In \emph{CVPR}, 2013.

\bibitem[Sun et~al.(2013)Sun, Wang, and Tang]{sun2013deep}
Yi~Sun, Xiaogang Wang, and Xiaoou Tang.
\newblock Deep convolutional network cascade for facial point detection.
\newblock In \emph{CVPR}, 2013.

\bibitem[Tsogkas et~al.(2015)Tsogkas, Kokkinos, Papandreou, and
  Vedaldi]{tsogkas2015semantic}
Stavros Tsogkas, Iasonas Kokkinos, George Papandreou, and Andrea Vedaldi.
\newblock Semantic part segmentation with deep learning.
\newblock \emph{arXiv preprint arXiv:1505.02438}, 2015.

\bibitem[van~den Oord et~al.(2016)van~den Oord, Kalchbrenner, and
  Kavukcuoglu]{oord2016pixel}
Aaron van~den Oord, Nal Kalchbrenner, and Koray Kavukcuoglu.
\newblock Pixel recurrent neural networks.
\newblock \emph{arXiv preprint arXiv:1601.06759}, 2016.

\bibitem[Visin et~al.(2015)Visin, Kastner, Cho, Matteucci, Courville, and
  Bengio]{visin2015renet}
Francesco Visin, Kyle Kastner, Kyunghyun Cho, Matteo Matteucci, Aaron
  Courville, and Yoshua Bengio.
\newblock Renet: A recurrent neural network based alternative to convolutional
  networks.
\newblock \emph{arXiv preprint arXiv:1505.00393}, 2015.

\bibitem[Williams and Zipser(1995)]{williams1995gradient}
Ronald~J Williams and David Zipser.
\newblock Gradient-based learning algorithms for recurrent networks and their
  computational complexity.
\newblock \emph{Back-propagation: Theory, architectures and applications},
  pages 433--486, 1995.

\bibitem[Yamashita et~al.(2015)Yamashita, Nakamura, Fukui, Yamauchi, and
  Fujiyoshi]{Yamashita2015}
Takayoshi Yamashita, Takaya Nakamura, Hiroshi Fukui, Yuji Yamauchi, and
  Hironobu Fujiyoshi.
\newblock Cost-alleviative learning for deep convolutional neural network-based
  facial part labeling.
\newblock \emph{IPSJ Transactions on Computer Vision and Applications}, 2015.
\newblock \doi{10.2197/ipsjtcva.7.99}.

\bibitem[Zheng et~al.(2015)Zheng, Jayasumana, Romera-Paredes, Vineet, Su, Du,
  Huang, and Torr]{crfasrnn_iccv2015}
Shuai Zheng, Sadeep Jayasumana, Bernardino Romera-Paredes, Vibhav Vineet,
  Zhizhong Su, Dalong Du, Chang Huang, and Philip Torr.
\newblock Conditional random fields as recurrent neural networks.
\newblock In \emph{ICCV}, 2015.

\bibitem[Zuo et~al.(2015)Zuo, Shuai, Wang, Liu, Wang, Wang, and
  Chen]{zuo2015convolutional}
Zhen Zuo, Bing Shuai, Gang Wang, Xiao Liu, Xingxing Wang, Bing Wang, and Yushi
  Chen.
\newblock Convolutional recurrent neural networks: Learning spatial
  dependencies for image representation.
\newblock In \emph{CVPR Workshops}, 2015.

\end{thebibliography}
}
\end{document}